\documentclass[twoside]{article}

\usepackage[accepted]{aistats2023}
\usepackage{hyperref}
\usepackage{url}
\usepackage{graphicx}
\usepackage{booktabs}
\usepackage{amsmath}
\usepackage{amssymb}
\usepackage{float}
\usepackage{pifont}
\usepackage{booktabs}
\usepackage{subcaption}
\usepackage{comment}
\usepackage{tcolorbox}
\usepackage{mdframed} 
\usepackage{xcolor}
\usepackage{algpseudocode}
\usepackage{xparse} 
\definecolor{graylight}{cmyk}{.30,0,0,.67} 

\newmdenv[ 
  linecolor=graylight,
  topline=false,
  bottomline=false,
  rightline=false,
  skipabove=\topsep,
  skipbelow=\topsep
]{leftrule}

\NewDocumentEnvironment{Example}{O{\textbf{Example:}}} 
{\begin{leftrule}\noindent\textcolor{graylight}{#1}\par}
{\end{leftrule}}

\usepackage{bm}

\usepackage{multirow}
\usepackage[ruled,vlined]{algorithm2e}
\usepackage{amsfonts, amsmath, microtype, hyperref, graphicx, booktabs, mdframed,enumitem}
\usepackage{xcolor}

\newcommand{\xmark}{\ding{55}}%

\DeclareMathOperator*{\argmin}{arg\,min}
\begin{document}

%

%

\twocolumn[

\aistatstitle{Retrospective Uncertainties for Deep Models using Vine Copulas}

\aistatsauthor{ Nataša Tagasovska \And Firat Ozdemir \And  Axel Brando }

\aistatsaddress{ Prescient Design, Genentech  \And  Swiss Data Science Center, \\  EPFL \& ETHZ \And Barcelona Supercomputing Center } ]

\begin{abstract}

Despite the major progress of deep models as learning machines, uncertainty estimation remains a major challenge.
Existing solutions rely on modified loss functions or architectural changes.
We propose to compensate for the lack of built-in uncertainty estimates by supplementing any network, retrospectively, with a subsequent vine copula model, in an overall compound we call Vine-Copula Neural Network (VCNN). 
Through synthetic and real-data experiments, we show that VCNNs could be task (regression/classification) and architecture (recurrent, fully connected) agnostic while providing reliable and better-calibrated uncertainty estimates, comparable to state-of-the-art built-in uncertainty solutions.
\end{abstract}

\begin{table*}[ht]
\small
    \caption{Overview of popular solutions that provide model and epistemic uncertainty for deep models within a unified framework.}
    \begin{center}
        \resizebox{0.95\textwidth}{!}{
            \begin{tabular}{lcccccc}
                \hline
                \hline
                \textbf{}                & \textbf{\begin{tabular}[c]{@{}c@{}}No custom \\ loss\end{tabular}} & \textbf{\begin{tabular}[c]{@{}c@{}}No  custom \\ architecture\end{tabular}} & \textbf{\begin{tabular}[c]{@{}c@{}}Single \\model\end{tabular}} & \textbf{\begin{tabular}[c]{@{}c@{}}Task\\  agnostic\end{tabular}} & \textbf{Scalable}       & \textbf{}	\textbf{\begin{tabular}[c]{@{}c@{}} Retrospective   \\ uncertainties\end{tabular}}\\
                \midrule
                \textbf{MC-dropout}         & \checkmark  & \xmark & \checkmark        & \checkmark & \checkmark & \xmark \\
                \textbf{Ensembles}        & \checkmark & \checkmark & \xmark    & \xmark  & \xmark   & \xmark   \\
                \textbf{Bayesian NNs}    & \xmark  & \xmark & \checkmark    & \checkmark  & \xmark & \xmark      \\
                \textbf{Vine-Copula NNs} & \checkmark      & \checkmark      &  \checkmark       & \checkmark & \checkmark\footnotemark  & \checkmark  \\ \hline 		\hline
        \end{tabular}}
        \label{tab:comparison}
    \end{center}
\end{table*}

\section{INTRODUCTION} \label{sec:intro}


Despite the high performance of deep models in recent years, industries still struggle to include neural networks (NNs) in production, at fully operational levels. 
Such issues originate in the lack of confidence estimates of deterministic neural networks, inherited by all architectures, convolutional, recurrent, and residual ones to name a few. 
Accordingly, a significant amount of effort has been put into making NNs more trustworthy and reliable \cite{lakshminarayanan2017simple, Gal2016Uncertainty, guo2017calibration, tagasovska2018single, brando2019umal, havasi2020training}. 
Given their high predictive performance, one would expect that a deep model could also provide reasonably good confidence estimates.
however, the challenge persists in how to yield these estimates without excessively disrupting the model, i.e. impacting its accuracy and train/inference time.

\begin{figure}[t]
    \centering
    \includegraphics[width=0.45\textwidth]{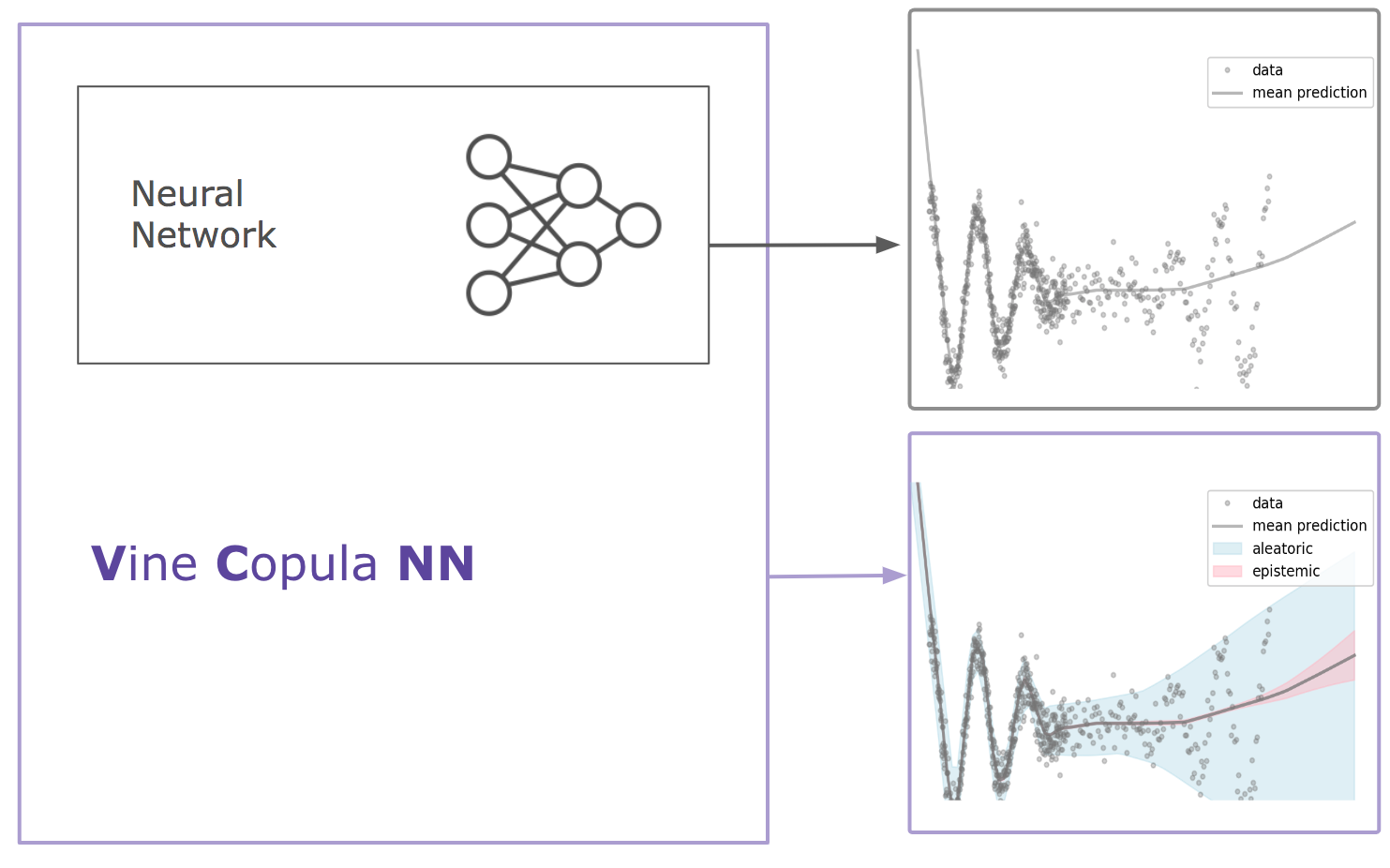}
    \caption{VCNN: We propose a plug-in vine-copula module that can complement any neural network with uncertainty estimates, any time after a model has been trained, without requiring any modifications to it. Additionally, our intervals capture both - aleatoric and epistemic uncertainty.}
    \label{fig:intro}
\end{figure}

Typically, a NN does not quantify uncertainty, but merely provides point estimates as predictions.
The predictive uncertainty, associated with the errors of a model, originates from two probabilistic sources:
\begin{itemize}
    \item an \emph{epistemic} one - the data provided in the training is not complete, i.e. certain input regions have not been covered in the training data, or the model lacks the capacity to approximate the true function (lack of knowledge, hence, reducible);
    \item an \emph{aleatoric} one, resulting from the inherent noise in the data, hence, an irreducible quantity, but, can be accounted for. 
\end{itemize}
To use NNs for applications in any domain, it is essential to provide uncertainty statements related to both of these sources. 
Nowadays, the most popular approaches for overall uncertainty estimation used by practitioners are MC-dropout \cite{gal2016dropout, kendall2017uncertainties}, ensembles \cite{lakshminarayanan2017simple} and Bayesian NNs \cite{hernandez2015probabilistic}.
Each of these approaches comes at a price, whether in terms of accuracy or computation time, as summarised in \autoref{tab:comparison}. 

This work is an attempt to alleviate those costs, reclaiming uncertainties for any deterministic model, \emph{retrospectively}, by complementing it with a \emph{vine copula} \cite{joe2011dependence}.
We favor vine copulas for the flexible estimation (parametric and non-parametric) as well as the (theoretically justified) scalability.

%

We use
\autoref{tab:comparison} to position the unique properties of having vine copula uncertainty estimates.
The vine copula provides an elegant way to extend any trained network retrospectively by quantifying both epistemic and aleatoric uncertainties, with performance on par or better than popular baselines. 
Given the undesired training costs of deeper and wider NNs, VCNN strikes as a viable solution.
which is particularly attractive given the increasing training costs (both financial and environmental) of NNs.  

It is also important to note that although there are many recent developments regarding the uncertainty of deep models, only a few consider the two sources (aleatoric and epistemic), and even fewer have a unifying framework for both of them. This is why, in sciences and industry which require both, Bayesian NNs prevail regardless of their heavy implementation.
Motivated by this practical problem, our contributions are as follows:
	\begin{itemize}
\setlength\itemsep{-0.5em}
		\item A new methodology for recovering uncertainties in deep models based on vine copulas (\autoref{sec:methodology}),
		\item An implementation of plug-in uncertainty estimates  (\autoref{alg:vcnn}), 
		\item Empirical evaluation on real-world datasets (\autoref{sec:experiments}).
	\end{itemize}

\section{BACKGROUND}
\label{sec:copula_uncertainties}

\textbf{Problem setup}
We consider a supervised learning setup where $X \in \mathbb{R}^{p}$ is a random variable for the features, and $Y \in \mathbb{R}$ is the target variable. 
We assume a process $y = f(x, \epsilon)$ to be responsible for generating a dataset $\mathcal{D} = \lbrace {x_i, y_i\rbrace}_{i=1}^N$ from which we observe realizations.
We are interested in learning a prediction model for $f: \mathcal{X} \rightarrow \mathcal{Y}$. In particular, to approximate $f$, we chose a (deep) neural network, with its weights considered as model parameters $\Theta$.
The DNN can learn a model that approximates the conditional mean well, i.e  $ \hat{f}_{\Theta} \in \argmin_{f}  \ell(\hat{f}_{\Theta}(x), y)$ by setting the loss $\ell$ to mean-squared error.

\begin{figure*}[t]
	\begin{center}
		\includegraphics[width=0.98\textwidth]{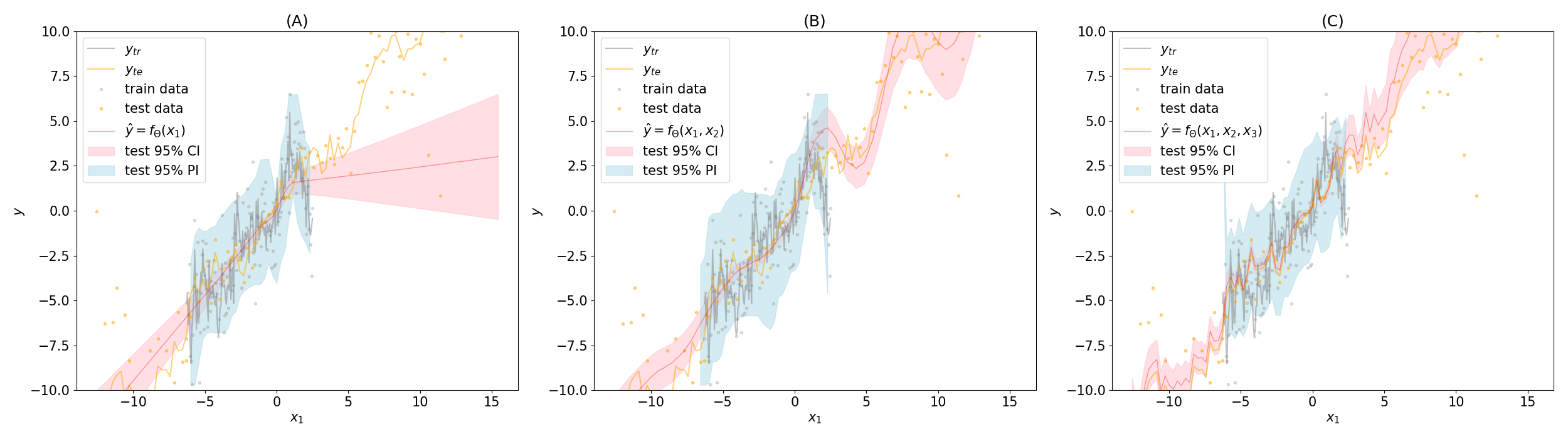}
	\end{center}
	\caption{Confidence and Prediction intervals by VCNN in a toy regression problem. Data generated as: $Y_{train} = X_1 + X_2 + X_3 + \epsilon$, with $X1 = \mathcal{U}[-2\pi, 2\pi]$, $X_2 = sin(2X_1)$, $X_3 \sim sin(X_1^2)$, with $\epsilon \sim \mathcal{N}(0, 0.2)$. For the test data $Y_{train} = X_1 + X_2 + X_3 + x_1\epsilon$, with $X1 \sim \mathcal{U}[-4\pi, 5\pi]$, $X_2 = sin(X_1)$, $X_3 \sim sin(X_1^2)$, with $\epsilon \sim \mathcal{N}(0, 0.5)$. 
Train data consists of 280 samples, and the test of 100. The network was trained for 100 epochs for the three presented cases.}
	\label{fig:motivating_example}
\end{figure*}

Within standard statistical models, one could relate the epistemic uncertainty to \emph{confidence intervals} (CIs) which evaluate the $Pr(f(x) | \hat{f}(x))$, whereas the aleatoric, noise uncertainty is captured by \emph{prediction intervals} (PIs) $Pr(y| \hat{f}(x))$. 
Both of these uncertainties contribute to the variability in the predictions which can be expressed through entropy or variance  \cite{lobato_decomposition}.
In this work, we will use the variance $\sigma^2$, as a quantity of interest unifying the two sources of uncertainty, and represent the overall predictive uncertainty per data point by its lower bound
$\displaystyle L := \min_{i \in N}  ({\sigma^2_{e^i}}, {\sigma^2_{a^i}})$ and its upper bound $\displaystyle U:= \max_{i \in N}  ({\sigma^2_{e^i}}, {\sigma^2_{a^i}})$, where $a$ stands for aleatoric (data related uncertainty) and $e$ for epistemic ( model uncertainly).
Our goal is to propose a tool that recovers estimates for \emph{both the confidence and the prediction intervals} of a DNN.  This natural decomposition wrt variance will further allow for them to be combined and used in applications as deemed most suitable\footnote{Different domains have different ways of adding up or considering the two uncertainties in final applications \cite{der2009aleatory}}. We summarize our method VCNN in \autoref{fig:intro}.

We propose \autoref{fig:motivating_example} to exemplify the sources and respective uncertainty estimates in more detail.
The observations are following a wiggly sequential data where the true generative process is a function $f(X_1, X_2, X_3, \epsilon)$ which we try to predict with a DNN of 2 hidden layers with 50 neurons each.
Moving from left to right, the plots in \autoref{fig:motivating_example} show the network predictions given: A) only observations from $X_1$ as input, i.e. $\hat{y} = \hat{f}_\Theta(x_1)$, then $\hat{y} = \hat{f}_\Theta(x_1, x_2)$ in B, and $\hat{y} = \hat{f}_\Theta(x_1, x_2, x_3)$ in C.
By including relevant variables (more knowledge; more data), the fitted model improves in each consecutive plot, visible also through its CI and PI.
This toy example shows the sensitivity of our estimates to both sources: As the DNN gets more information, 
the prediction of the model improves, its CIs improve as well 
(narrower in train region, wider outside), while the PIs capture the noisiness of the data (when variables are omitted, such as in A and B, this translates into more stochasticity, i.e as aleatoric noise). 
Moreover, from \autoref{fig:motivating_example}, we also see why considering both sources is important: in the out-of-distribution region, the confidence intervals prevail, while in the nosier regions, the prediction intervals envelop the CI. %

Next, we present how we envision using the natural properties of vine-copulas to account for the uncertainties of a deep model.
We assume that a DNN, $ \hat{f}_{\Theta}$, has already been trained until convergence for a specific task. 
Our goal is to supplement the output of the already-trained network, with trustworthy uncertainty estimates.
For efficiency and simplicity of the method, we use the embeddings from the last hidden layer of the network - $\xi$, (where $\xi = h(x) \in \mathbb{R}^d, d << p$) as a proxy for the overall network parameters. 

\subsection{Copulas and Vines} 
\begin{figure}[t]
    \centering
    \includegraphics[width=0.45\textwidth]{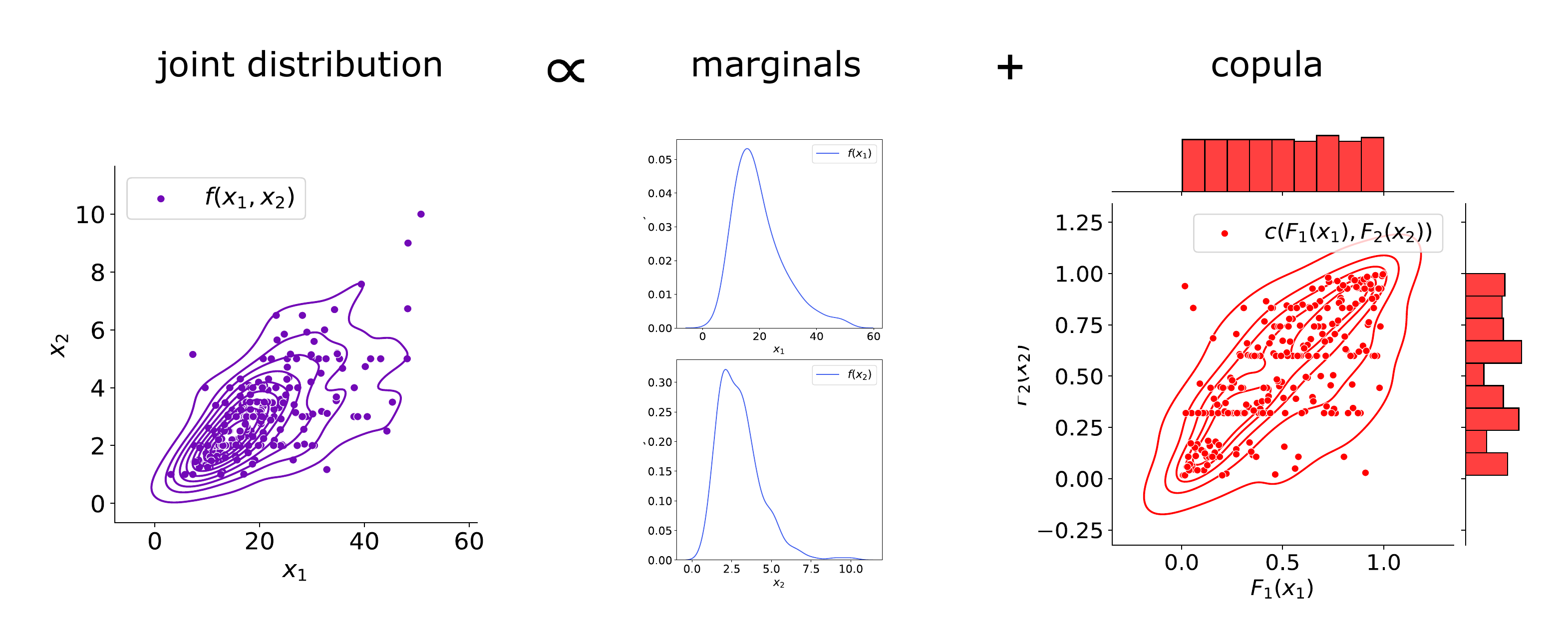}
    \caption{Expressing joint densities with copulas.}
    \label{fig:copula}
\end{figure}

According to  Sklar’s theorem \cite{sklar1959}, the joint density of any bivariate random vector $(X_1, X_2)$, can be expressed as
\begin{align}\label{eq:cop}
    f(x_1, x_2) = f_1(x_1)f_2(x_2)c(F_1(x_1), F_2(x_2))
\end{align}
where $f_i$\footnote{In this section, we use the standard notation for densities ($f$) and distributions ($F$) as in the copula literature} are the marginal densities, $F_i$ the marginal distributions, and $c$ the copula density. 
That is, any bivariate density is uniquely described by the product of its marginal densities and a \emph{copula density}, which is interpreted as the \emph{dependence structure}. \autoref{fig:copula} illustrates all of the components representing the joint density. 
As a benefit of such factorization, by taking the logarithm  on both sides, one could straightforwardly estimate the joint density in two steps, first for the marginal distributions, and then for the copula. 
Copulas are widely used in finance and operation research, and, in the machine learning community have recently been integrated with deep models for generative and density estimation purposes \cite{tagasovska2019copulas, janke2021implicit, drouin2022tactis, hassan2022}.
Hence, copulas provide means to flexibly specify the marginal and joint distribution of variables. For further details, please refer to \cite{aas2009pair, Joe2010}.

\begin{figure}[th!]
    \centering
    \includegraphics[width=0.48\textwidth]{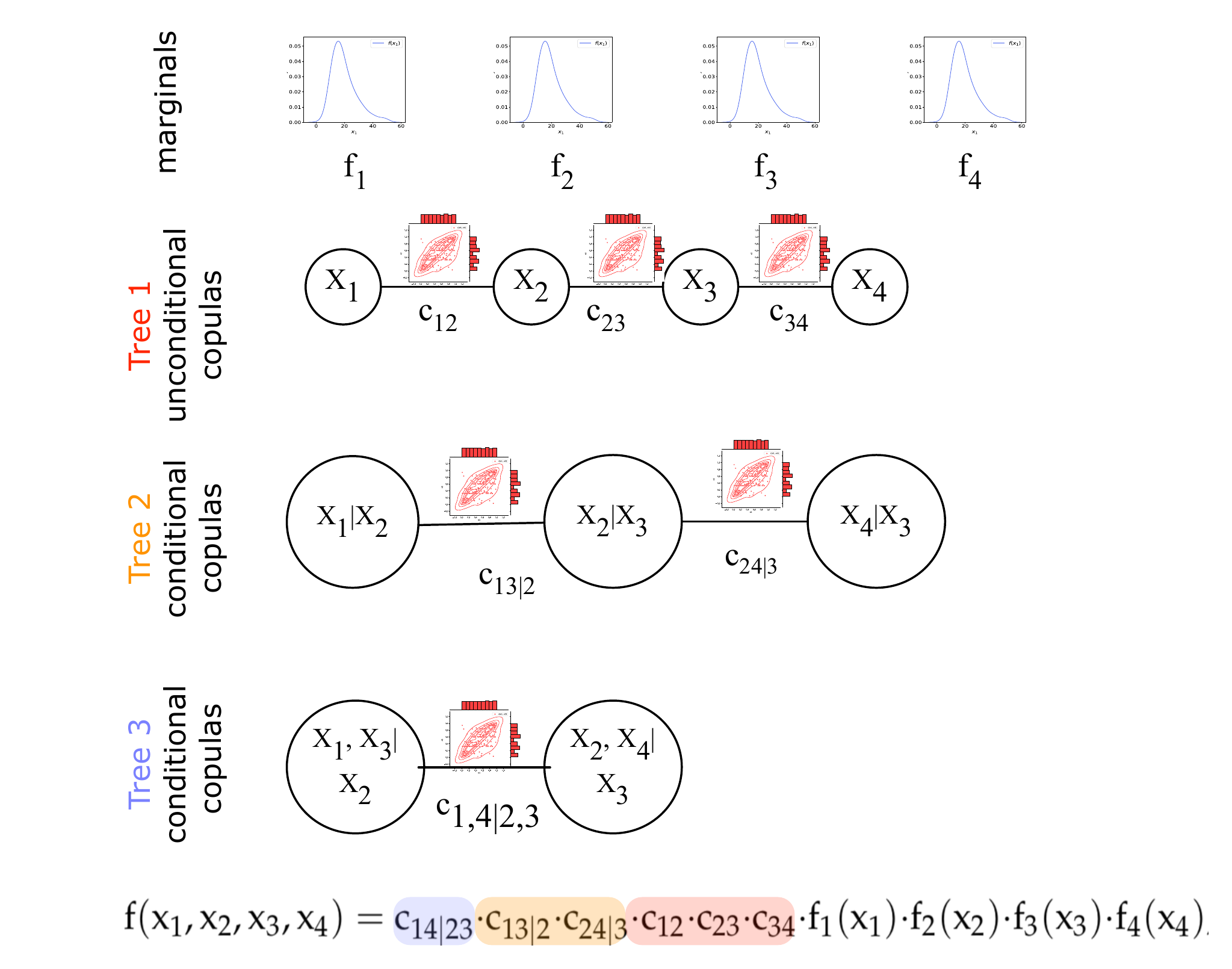}
    \caption{Multivariate joint density factorized with a vine copula.}
    \label{fig:vine}
\end{figure}

There exist many parametric representations through different copula families, however, to leverage even more flexibility, in this paper, we focus on kernel-based nonparametric copulas of \cite{Geenens2017}.
\autoref{eq:cop} can be generalized and holds for any number of variables.
To be able to fit densities of more than two variables, we make use of the pair copula constructions, namely \emph{vines}; hierarchical models, constructed from cascades of bivariate copula blocks \cite{nagler2017nonparametric}.
According to \cite{Joe97,Bedford02}, any $d$ dimensional copula density can be decomposed into a product of $\frac{d(d - 1)}{2}$ bivariate (conditional) copula densities.
Although such factorization may not be unique, it can be organized in a graphical model, as a sequence of $d - 1$ nested trees, called \emph{vines}.
We denote a tree as $T_m = (V_m, E_m)$ with $V_m$ and $E_m$ the sets of nodes and edges of tree $m$ for $m = 1, \dots, d-1$. Each edge $e$ is associated with a bivariate copula. An example of a vine copula decomposition is given \autoref{fig:vine}.

In practice, in order to construct a vine, one has to choose two components: 
\begin{enumerate}
    \item the structure, the set of trees $T_m = (V_m, E_m)$ for $m = 1, \dots, d-1$
    \item  the pair-copulas, the models for $ c_{j_e, k_e | D_e}$ for $e \in E_m$ and $m = 1, \dots, d-1$.
\end{enumerate}

Corresponding algorithms exist for both of those steps and in the rest of the paper, we assume consistency of the vine copula estimators for which we use the implementation by \cite{Nagler2018}. 

\section{VINE-COPULA UNCERTAINTY ESTIMATES}
\label{sec:methodology}

\subsection{Simulation-based Confidence Intervals}
\label{sec:vc_ci}
Besides being a flexible density estimation method, (vine) copulas additionally have generative properties \cite{Dissmann2013}. 
That means that once a copula is fit on a data distribution, it can be used to produce random observations from it.
Although mainly used for stress testing in the finance domain, this property of copulas has recently been recognized as a useful high dimensional generative model for images \cite{tagasovska2019copulas}.
Hence, this served as an inspiration to use vine copulas to bootstrap the network parameters, yielding confidence intervals for its predictions via simulations.
%
Simulation-based approaches for confidence intervals have previously been used in the literature for smoothing splines \cite{ruppert2003semiparametric}.
%
	
Following a similar approach to \cite{ruppert2003semiparametric}, we consider
the true function $f$ over $L$ locations in $x$, $\mathbf{l} = \lbrace{l_1, l_2, \dots l_L \rbrace}$, $f_\mathbf{l}$ denoting the vector of evaluations of $f$ at each of those locations, and the corresponding estimate of the true function by the trained DNN as $\hat{f}_{\Theta_\mathbf{l}}$.
The difference between the true function and our unbiased\footnote{We consider a network of sufficient capacity such that it can approximate any function (considering NNs as universal approximators~\cite{hornik1991approximation}), hence, we exclude the model to true function bias and consider only the variability of the models' parameters with respect to the data.} estimator is given by:
\begin{align*}
    \hat{f}_{\Theta_\mathbf{l}} - f_\mathbf{l} = H_\mathbf{l}\begin{bmatrix} \hat{\Theta} - \Theta\ \end{bmatrix} 
\end{align*}

where $H_\mathbf{l}$ is the evaluation of the DNN at the locations $\mathbf{l}$, and the expression in brackets represents the variation in the estimated network parameters.
The distribution of the variation is unknown, and we aim to approximate it by simulation.:
%
A $100(1 - \alpha)\%$ simultaneous confidence interval is:
\begin{align}
\hat{f_l} \pm q_{1-\alpha} 
	\hat{\sigma} [(\hat{f}(l_j) - f(l_j)]_{j=1}^L
\end{align}
where $\sigma$ denotes a standard deviation and $q_{1-\alpha}$ is the $1 - \alpha$ quantile of the random variable:
\begin{align}
  	\sup_{x \in X} \begin{vmatrix}
		\frac{\hat{f}(x)- f(x)}
		{\hat{\sigma}(\hat{f}(x)) - f(x)}
	\end{vmatrix} \approx \max_{1 \leq j \leq L} \begin{vmatrix} 
		\frac{H_l\begin{bmatrix} \hat{\Theta} - \Theta\\ \end{bmatrix} _j}
		{\hat{\sigma}\hat{f}(l_j) - f(l_j))}
	\end{vmatrix}. 
\end{align}

The $\sup$ refers to the supremum or the least upper bound; which is the least value of $x$ from the set of all values which we observed, which is greater than all other values in the subset. 
Commonly, this is the maximum value of the subset, as indicated by the right-hand side of the equation. Hence we want the maximum (absolute) value of the ratio overall values in $\mathbf{l}$.
The fractions in both sides of the equation correspond to the standardized deviation between the true function and the model estimate, and we consider the maximum absolute standardized deviation. 
While we do not have access to the distribution of the deviations, we only need its quantiles that we can approximate via simulations. 

Herein, we exploit the generative nature of copulas whereby we ``bootstrap" the trained DNN with the vine copula fitted over the embeddings $\xi = h(x)$ (the embeddings of a data point $x$ in the last hidden layer) and target $y$.
For each simulation, we find the maximum deviation of the (re)-fitted functions from the true function over the grid of $\mathbf{l}$ values we are considering. 

\footnotetext{Vine Copulas are theoretically designed to deal with high-dimensional spaces \cite{nagler2019model}.}

%
%
As the last step, we find the critical value to scale the standard errors $SE$ such that they yield the simulation interval; we calculate the critical value $r$ for a 95\% simulation confidence interval using the empirical quantile of the ranked standard errors. 
Finally, we recover the upper and lower epistemic bounds obtained as:

\begin{equation}\label{eq:epi}
    \tcbox[nobeforeafter]{\( y_{Ue}/y_{Le} = \hat{f}_\Theta (x) \pm r*SE(H_l\begin{bmatrix} \hat{\Theta} - \Theta\ \end{bmatrix}) \)}.
\end{equation}
%


\subsection{Conditional-quantile Prediction Intervals}

In the case of predictive aleatoric uncertainty, we are interested in estimating the conditional distribution of the target variable given the inputs $F(Y|X)$, which is captured by conditional quantiles of interest \cite{tagasovska2018single}. 
Since we are aiming for a unified framework for uncertainties, we wish to also derive the necessary quantile predictions from vine-copula representations. To do so, we use the recent development by \cite{nagler2018solving} which shows that conditional expectations can be replaced by unconditional ones, and such property can be used to leverage copulas for solving regression problems. This approach allows for new estimators of various nature, such as mean, quantile, expectile, exponential family, and instrumental variable regression, all based on vine-copulas.
Regression problems can be represented as solving equations:
\begin{align*}
    \mathbb{E}\lbrace g_\phi(Y) | X \rbrace = 0
\end{align*}
where $\phi$ is a parameter of interest and the set $\mathcal{G} = \lbrace g_\phi: \phi \in \Phi \rbrace$ is a family of identifying functions. Conditional expectations can be replaced by the less challenging to estimate, unconditional ones as follows: 
\begin{align*}
      \mathbb{E}\lbrace g_\phi(Y) | X \rbrace  = \mathbb{E}\lbrace g_\phi(Y) w^*(Y) \rbrace 
\end{align*}

with 
\begin{align}\label{eq:weights}
    w^*(y) = \frac{dF_{Y|X}(y|x)}{dF_Y(y)} = \frac{dF_{Y,X}(Y, X)}{dF_X(x)dF_Y(y)}.
\end{align}
Given the background on copulas, we see that the weights in \autoref{eq:weights} can be expressed with:
\begin{align*}
     w^*(y) = \frac{dC_{Y,X}(F_Y(y), F_X(x))}{dF_{C_X}(F_X(x))dF_{C_Y}(F_Y(y))}.
\end{align*}
This expression can be further simplified by dropping $C_X$ as it does not depend on $y$ or $\phi$, and because copulas have uniform margins; we are left with $w^*(y) = dC_{Y,X}(F_Y(y), F_X(x))$.
As exemplified in \cite{nagler2018solving}, with different identifying functions $g_\phi$, we get estimators $\hat{\phi} = \hat{\phi}(Y_1, X_1, \dots, Y_n, X_n)$ to various conditional distributions as solutions to $ \frac{1}{n} \sum_{i=1}^N g_\phi(Y_i)w(Y_i) = 0$ .
Here, we rely on solving estimating equations when the identifying function $g_\phi$ is suitable for quantile regression.
We thus let:
\begin{align*}
    \phi_{\tau} = F^{-1}_{Y|X}(\tau | X = x)
\end{align*}
be the conditional $\tau-$quantile
for $\tau \in (0, 1)$, all levels considered jointly. 
Using this knowledge, and the embeddings $\xi$ as conditioning variables, we are now able to compute the upper and lower bound of a required $(1 - \alpha)$ prediction interval, by obtaining:


\begin{equation}\label{eq:ale}
    \tcbox[nobeforeafter]{\( y_{Ua} = F^{-1}_{Y|\xi}((1 - \frac{\alpha}{2})|\xi)  \quad  y_{La} = F^{-1}_{Y|\xi}((\frac{\alpha}{2})|\xi) \\
\)}
\end{equation}
solving this conditional expectation using vine copulas. 
In practice, we use the recent implementation from the \verb|eecop| package \cite{Nagler2020}.
%

With the expressions capturing both uncertainty estimates, we finally  summarize our method VCNN, computing both epistemic and aleatoric uncertainties per data point in Algorithm~\ref{alg:vcnn}.

\begin{algorithm}
\SetAlgoLined
\textbf{Inputs:} DNN $\hat{f}_\Theta$, train samples $\lbrace{ x_i, y_i \rbrace}_{i=1}^N$, test samples $\lbrace{x_j^{\prime} \rbrace}_{j=1}^L$\;
Step 1. Obtain embeddings from the last dense layer, before the activations, 
		$\xi = \lbrace {h(x_i)\rbrace}_{i=1}^N \in \mathbb{R}^d$\;
Step 2. Fit a vine copula with the embeddings and target variable, $V:=C(\xi_1, \xi_2, \dots \xi_d, y)$. \;
\textbf{Epistemic uncertainty} \;
Step 3.  For $S$ repetitions:
		\begin{enumerate}
			\item Sample random observations from the vine $V$, $\lbrace{ \xi^*,  y^* \rbrace}$.
			\item Re-train the last hidden and the output layer of $\hat{f}_\Theta$ with $\lbrace{ \xi^*,  y^*\rbrace}$ .
			\item Save the bootstraped $S$ heads\footnotemark of the network.
		\end{enumerate}
Step 4. Compute the confidence interval (CI) bounds: $y_{Le}(x^{\prime})$ and $y_{Ue}(x^{\prime})$ with \autoref{eq:epi}. \;
\textbf{Aleatoric uncertainty} \;
Step 5. Compute the prediction interval (PI) bounds: $y_{La}(x_j^{\prime})$ and $y_{Ua}(x_j^{\prime})$ by solving for the required quantiles $F^{-1}_{1 - \frac{\alpha}{2}}(Y|\xi(x^{\prime}))$ and $F^{-1}_{\frac{\alpha}{2}} (Y|\xi(x^{\prime}))$ using  vine $V$ with \autoref{eq:ale}.\\
 \textbf{Outputs} $S$-heads, vine copula $V$, confidence and prediction intervals for $x^{\prime}$.
 \caption{Vine-copula DNN uncertainty estimates.}\label{alg:vcnn}
\end{algorithm}
\footnotetext{We use the term ``heads" here as our bootstrapped layers are similar in architecture to the concept of multi-head networks.}

\section{EXPERIMENTS}
\label{sec:experiments}

 \begin{figure*}[th]
		\begin{center}
			\includegraphics[width=0.95\textwidth]{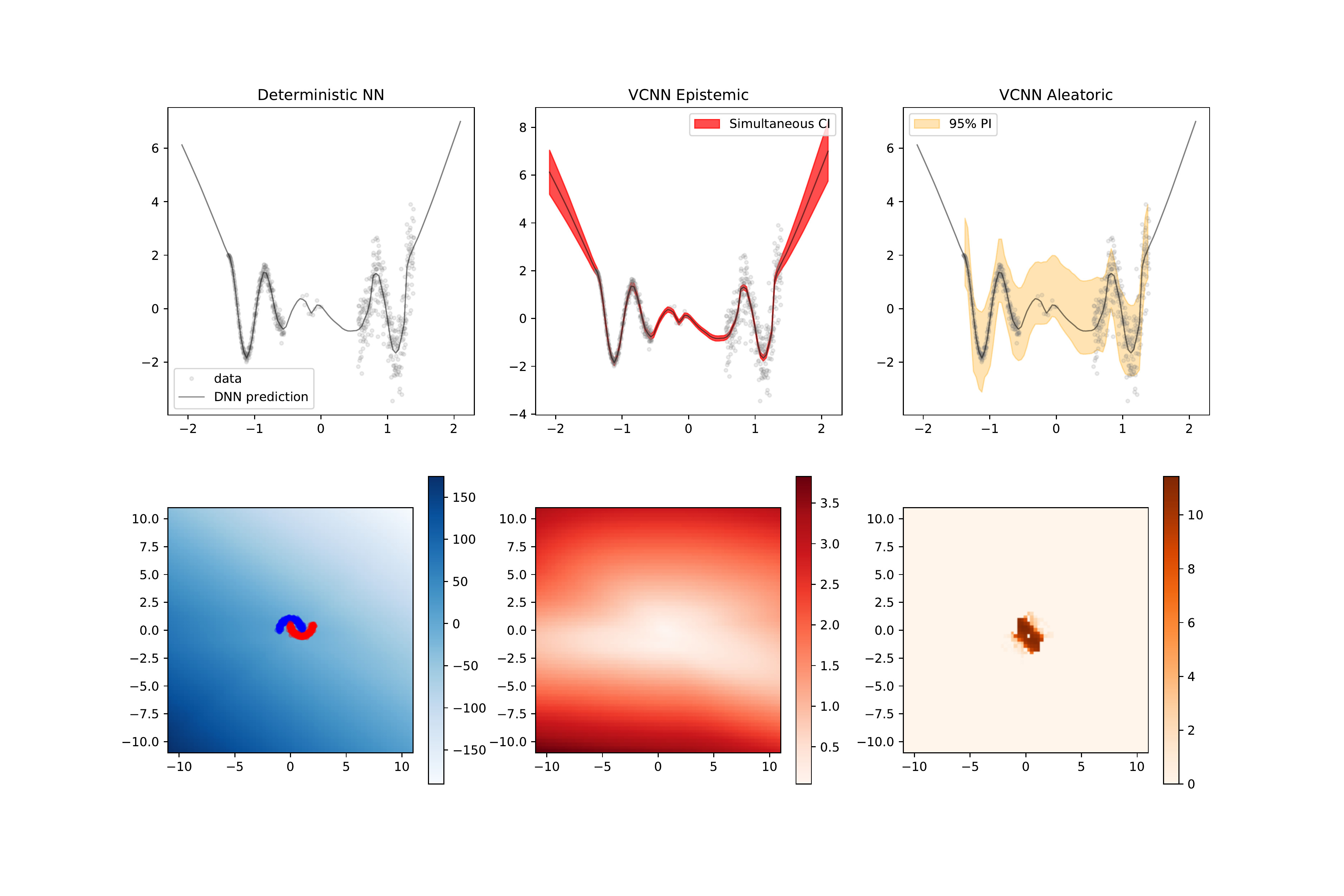}
		\end{center}
		\caption{VCNN for toy regression and classification scenarios.}
		\label{fig:motivating_example2}
\end{figure*}

To empirically evaluate our method we propose three setups: synthetic toy examples for regression and classification, and, two real-world datasets: Datalakes - estimating the surface temperature of Lake Geneva based on sensor measurements, hydrological simulations and satellite imagery, and an AirBnB apartment price forecasting dataset. 
In all cases we wish to estimate the predictive uncertainty, hence, we account for both epistemic and aleatoric uncertainty sources.
As baselines, we use Deep Ensembles \cite{lakshminarayanan2017simple}, MC-dropout \cite{kendall2017uncertainties} and Bayesian NN \cite{hernandez2015probabilistic},
all adjusted to capture the two types of uncertainties, simultaneously. 
Our code and demo for VCNN are available on \href{https://github.com/tagas/vcnn}{https://github.com/tagas/vcnn}.
\paragraph{Metrics} To compare baselines we use the number of points captured by the prediction intervals (PI), which
should capture some desired proportion of the observations,
$(1 - \alpha)$. 
\begin{align*}
    Pr(\hat{y}_{L^i} \leq y_i \leq \hat{y}_{U^i} ) = (1 - \alpha)
\end{align*}
In all experiments except if stated otherwise, we target the common choice $\alpha=0.05$. To evaluate the quality of the PIs, similarly to \cite{Pearce2018}, we denote a vector, $\mathbf{k}$, of length $n$ which represents whether each data point has been captured by the estimated PIs, with  NO $k_i \in [0, 1]$ given by,
\begin{align*}
    k_i = \begin{cases} 
    1, & if \quad y_{L^i} \leq y_i \leq y_{U^i} \\
    0, & otherwise.
    \end{cases} 
\end{align*}
We denote the total number of points captured as $c = \sum_{i=1}^n k_i$. With this, we can now define the two main metrics we will use: 
\begin{itemize}
    \item The Prediction Interval Coverage Probability - $PICP := \frac{c}{n}, $ - which indicates how many of the observations are captured inside the estimated interval. A coverage closer to the desired proportion (set by $\alpha$) is preferred;

    \item The Mean Prediction Interval Width -  \[MPIW:= \frac{1}{n} \sum_{i=1}^n \hat{y}_{U^i} - \hat{y}_{L^i} , \] - which indicates how tight are the predicted intervals, where lower values are better. 
\end{itemize}


\paragraph{Baselines} In general, the proposed baselines (Ensembles and MC Dropout) capture the epistemic and aleatoric uncertainty jointly. Regarding the aleatoric uncertainty, these baselines consider DNN  with outputs being the conditional parameters of a normal distribution, $\mathcal{N}(\mu_{\Theta}, \sigma_{\Theta})$. On the other hand, each baseline models the epistemic uncertainty in a different way (due to random initialisation).

For the sake of a fair comparison, we let both baseline functions, $\mathcal{N}(\mu_{\Theta}, \sigma_{\Theta})$ have as input the embedding of the last dense layer, $\xi$, \footnote{This decision was made after verifying empirically that the results were similar considering $x$ or $\xi$ as input of the baseline functions.} similar to Algorithm~\ref{alg:vcnn}.





\paragraph{Heteroscedastic Deep Ensemble}
\emph{(Ensemble+N)} consists of a set of $S$ functions pairs, $ \{ (\mu_{\Theta,i}, \sigma_{\Theta,i} )\}_{i=1}^S$, such that each pair is trained (1) on a different split from the training data set and (2) with different parameters initialization, in order to maximize the diversity between the individual pairs (which also includes the bootstrap\cite{ganaie2021ensemble}).





\paragraph{Heteroscedastic Monte-Carlo Dropout}
\emph{(MCDrop+N)} outputs the pair, $ \{ (\mu_{\Theta,i}, \sigma_{\Theta,i} )\}_{i=1}^S$, where each $i$ corresponds to a different, randomly selected connection, implemented as dropout layers. Differently than the Ensemble+N, here we considered the whole training set as proposed in \cite{gal2016dropout}.

\paragraph{Bayesian Neural Network}\emph{(Bayesian NN+N)} outputs the pair, $ \{ (\mu_{\Theta,i}, \sigma_{\Theta,i} )\}_{i=1}^S$, where each $i$ corresponds to a different, randomly selected sample from the random variables associated to all neuron weights. In particular, this random variable has a prior and posterior normal distribution optimized via variational inference \cite{graves2011practical}.

%
\subsection{Toy example}
In \autoref{fig:motivating_example2} we include the evaluation of VCNN on toy examples. The first row  is a one-dimensional input bi-modal regression task, while the second row is a classification task for a two-dimensional input, namely the moons dataset. 
We show the results from a deterministic DNN (predictions and class probabilities respectively) in the first column, the epistemic uncertainty estimate in the second, and the aleatoric uncertainty estimate in the third obtained with vine copulas.
For the classification task, since the input is two-dimensional, we present the uncertainty scores through a color range, depicting the distance between the upper and lower bounds of the corresponding CI or PI.
We consider these classification results encouraging, since both the epistemic uncertainty grows further away from the train data, and the aleatoric uncertainty is high only in the region where there is an overlap between the two classes, as expected.

\subsection{Datalakes}
Datalakes\footnote{\url{https://www.datalakes-eawag.ch}} tackles certain data-driven problems such as estimating lake surface temperature given a range of sensory, satellite imagery, and simulation-based datasets. 
One such dataset is of Lake Geneva where a sparse hourly dataset is collected between years 2018 to 2020.
Given the temporal nature of the observations, the model that provided the best prediction was a bidirectional (Bi)\cite{schuster1997bidirectional} long short-term memory (LSTM)\cite{Hochreiter1997LongSM} network.
%
Originally, this method uses MC-dropout in addition to a negative log-likelihood loss as in \cite{kendall2017uncertainties}.
We replace those uncertainties with the vine based ones and we present our results in \autoref{tab:datalakes}. 
Comparisons with ensemble methods were omitted due to both computational restrictions and the MC-dropouts method being a comparable proxy. 
From \autoref{tab:datalakes} we see that the VCNN achieves on-par estimates for PICP with (significantly) narrower interval widths as suggested by the MPIW. 
The measuring unit is Celsius degrees.

\begin{table}[H]
\caption{PICP and MPIW values for VCNN and baselines for the real-world datalakes dataset. Note that the very long training of the biLSTM network did not allow for multiple executions for obtaining std variations or results from ensembles.} \label{tab:datalakes}
\begin{center}
\begin{tabular}{@{}lllll@{}}
    \toprule
    \multirow{2}{*}{\textbf{BiLSTM}} & \multicolumn{2}{c}{\textbf{Train}}                                    & \multicolumn{2}{c}{\textbf{Test}}                                     \\ \cmidrule(l){2-5} 
    & \multicolumn{1}{c}{\textbf{PICP}} & \multicolumn{1}{c}{\textbf{MPIW}} & \multicolumn{1}{c}{\textbf{PICP}} & \multicolumn{1}{c}{\textbf{MPIW}} \\ \midrule
    \textbf{VCNN}                    & .97                              & 5.71                              & .85                              & \textbf{5.84}                             \\
    \textbf{MC-Dropout+N}              & .94                              & 6.97                              & \textbf{.86}                              & 6.88                              \\ \bottomrule
\end{tabular}
\end{center}
\end{table}\label{tab:datalakes}

\begin{figure*}[t]
	\begin{center}
		\includegraphics[width=0.9\textwidth]{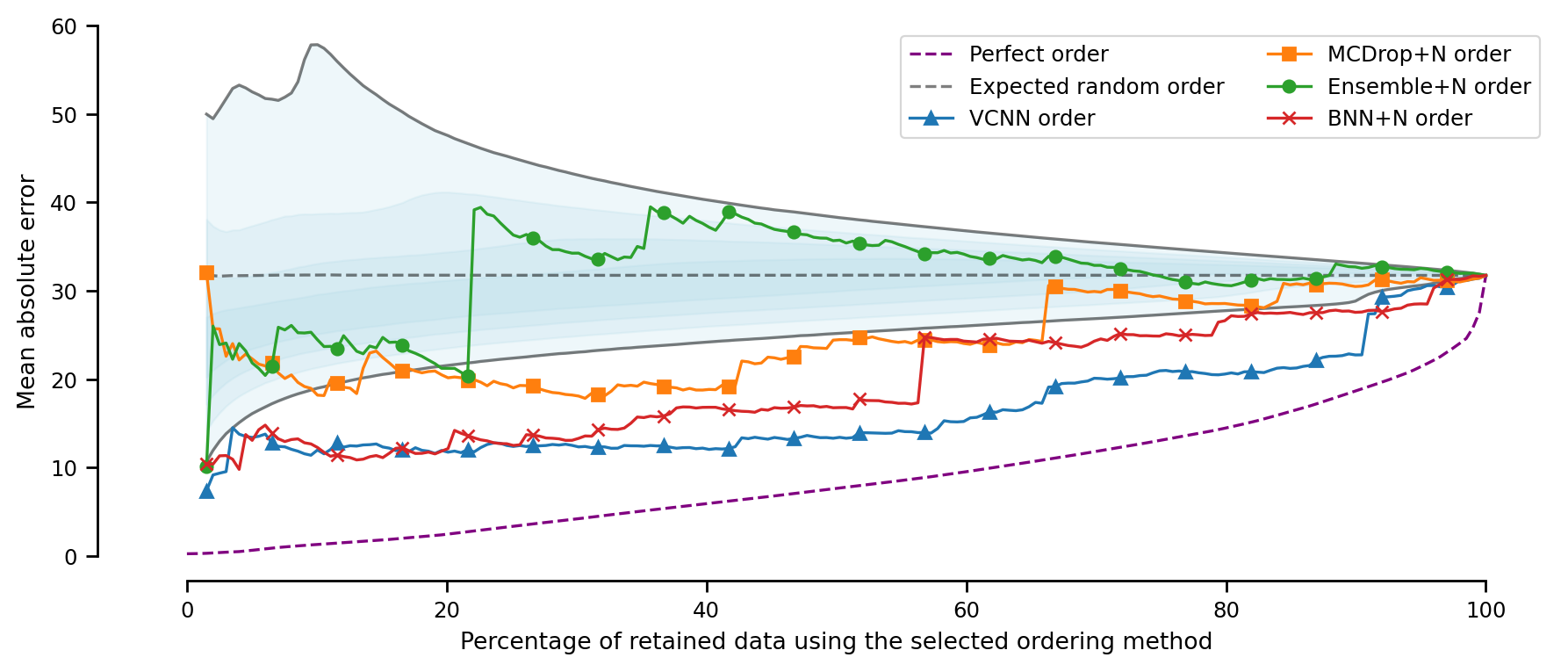}
	\end{center}
	\caption{Error-retention curves - considering the mean cumulative error for the most confident points and sorting using the uncertainty scores -, which are produced by the baselines and the VCNN (see \cite{brando2018uncertainty, brando2022t})}.
\label{fig:CAE}
\end{figure*}

\begin{figure}[h]
	\begin{center}
		\includegraphics[width=0.45\textwidth]{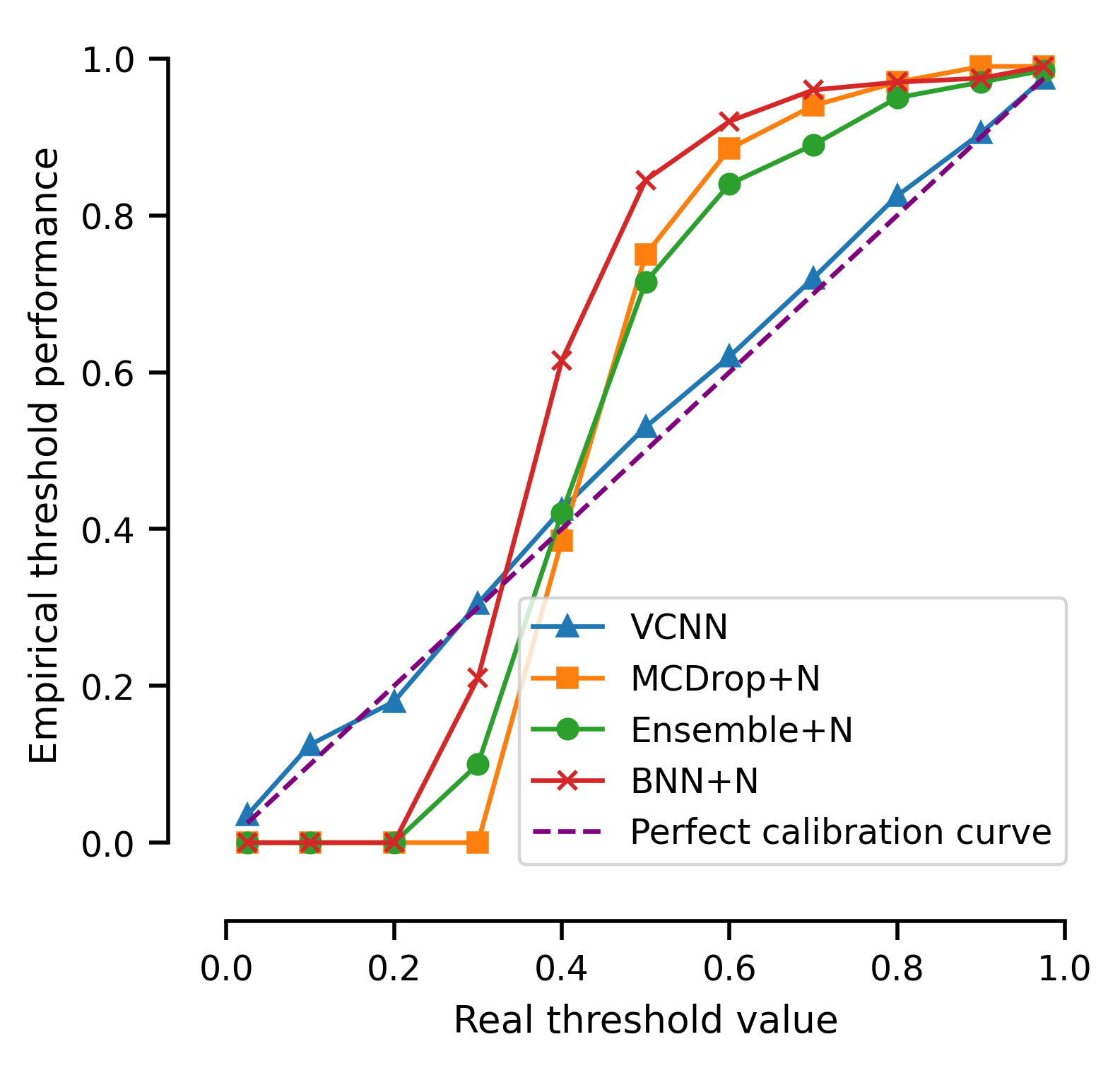}
	\end{center}
	\caption{Calibration curves of the baselines and the VCNN using their forecasted quantiles (see \cite{brando2022deep}).} 
\label{fig:calibration}
\end{figure}

\subsection{Room price forecasting} 


Based on publicly available data from the Inside Airbnb platform \cite{cox2019inside}, for Barcelona, we followed a regression problem proposed in \cite{brando2019umal}. The aim is to predict the price per night for $36,367$ flats using data from April 2018 to March 2019. The architecture used is a DenseNet, detailed in the Supplementary Material section.
From Table \ref{tab:airbnb} we notice that the plug-in estimates of VCNN outperform the Bayesian NNs, ensembles and MC-dropout - VCNN intervals are properly calibrated (targeting the 95\%) and they are tighter by at least a half. The measurement unit here is euros.


\begin{table}[H]
\caption{PICP and MPIW (mean and std. dev) for VCNN and baselines for the real-world Airbnb dataset.} \label{tab:airbnb}
\begin{center}
\resizebox{\linewidth}{!}{
\begin{tabular}{@{}lllll@{}}
    \toprule
    \multirow{2}{*}{\textbf{DenseNet}} & \multicolumn{2}{c}{\textbf{Train}}                                    & \multicolumn{2}{c}{\textbf{Test}}                                     \\ \cmidrule(l){2-5} 
    & \multicolumn{1}{c}{\textbf{PICP}} & \multicolumn{1}{c}{\textbf{MPIW}} & \multicolumn{1}{c}{\textbf{PICP}} & \multicolumn{1}{c}{\textbf{MPIW}} \\ \midrule
    \textbf{Bayesian NN+N}         &          $.99 \pm .00$          &         $698.45 \pm 93.$                               &                    $.99 \pm .00$          &               $645.55 \pm 73.$               \\
    \textbf{MC-Dropout+N}         &          $.99 \pm .00$          &         $468.03 \pm 49.$                               &                    $.99 \pm .00$          &               $514.15 \pm 72.$               \\
    \textbf{Ensemble+N}              &           $.99 \pm .00$                  &          $433.25 \pm 27.$                     &                     $.98 \pm .00$         &             $468.95 \pm 31.$                 \\
    \textbf{VCNN}                    &            $\boldsymbol{.97 \pm .01}$                &           $\boldsymbol{226.8 \pm 4.4}$                  &              $\boldsymbol{.95 \pm .00}$            &             $\boldsymbol{191.33 \pm 3.0}$                 \\
    \bottomrule
\end{tabular}
}
\end{center}
\end{table}
 \subsection{Evaluating the quality of uncertainty intervals}

Additionally, we can evaluate the quality of the reported uncertainty by leveraging this information as a confidence score. To accomplish this, we analyze an error-retention curve \cite{brando2022t}. This curve involves computing the normalized cumulative absolute errors in ascending order of their corresponding uncertainty scores, such that the individual error with the highest uncertainty score with respect to the selected ordering method appears at the end of the list. Moreover, we observe 
 in \autoref{fig:CAE}, that the VCNN exhibits an error-retention curve that consistently approximates the perfect curve, to a greater extent than other baselines. This plot reads as follows: with a perfect confidence score, we can retain 80\% of the data points and expect an MAE of 14 euros. For the same amount of data, with VCNN we expect MAE of 18 euros, and with all other baselines MAE $> 25$.



Finally, since for the AirBnB dataset all baselines can be trained in a reasonable time (including ensembles), we show a calibration plot in \autoref{fig:calibration}. Calibration lines are crucial to ensure proper behavior of the predicted quantiles, by plotting the model's predictions against the empirical quantile values. The results in \autoref{fig:calibration} show that VCNN forecasts are closer to the perfect calibration line while Ensembles, MC dropout, and Bayesian NNs underestimate quantiles lower than the median and are overconfident for quantiles above 0.5.





\subsection{Practical considerations for the vine copulas}

\textbf{Complexity} The complexity for fitting the vine copulas as currently implemented is approximate $ O(n \times dim \times vine depth)$ for estimation/sampling algorithms, both involving a double loop over-dimension/truncation level with an internal step scaling linearly with the sample size.
Due to this linear scaling in the number of samples and dimensions, we found it useful to randomly subsample the train data and use truncated vines. The runtimes of VCNN could greatly benefit of implementation optimizations, however, this is outside of the scope of the current work.

\paragraph{Hyperparameters} Although the vine structure and the copula family could be considered as hyperparameters, empirically we observed that the best results are obtained when we use nonparametric family with a low multiplier for the kernel width, i.e. 0.1, for both the confidence and prediction intervals. 
For computational purposes, we also chose to truncate the vine, that is to fit only the first 2 - 5 trees in the vine and consider all the subsequent ones as independent copulas. 
These factors can indeed appear as more important for other datasets and we are conducting multiple ablation studies to explore this further.

\section{CONCLUSION}
In this work, we present new, plug-in uncertainty estimates for neural networks using vine copulas which can be applied to any network retrospectively. 
Importantly, our estimates do not impact the performance of the original model in any way, while they manage to enhance it with faithful predictive confidence measures. 
%
%
This is particularly attractive given the increasing training costs (financial and environmental) with deeper and wider NNs.

We hope our method can help in increasing the trustworthiness of deep models, specifically for real-world scenarios which render critical decision-making. 
There are multiple directions for extensions: out-of-distribution detection methods in classification tasks underlying different architectures (image, graph), and, we expect VCNN to improve results in other tasks relying on uncertainties, such as segmentation or active learning. 
%

\section*{Acknowledgments}
The research leading to these results has received funding from the Horizon Europe Programme under the SAFEXPLAIN Project (www.safexplain.eu), grant agreement num. 101069595 and the European Research Council (ERC) under the European Union’s Horizon 2020 research and innovation programme (grant agreement No. 772773). Additionally, this work has been partially supported by Grant PID2019-107255GB-C21 funded by MCIN/AEI/ 10.13039/501100011033.

\bibliography{main}

$ $

\onecolumn
\section{SUPPLEMENTARY MATERIAL}

\subsection{Example of a Vine Copula}
A sequence is a vine if it satisfies the set of conditions which guarantee that the decomposition represents a \emph{valid joint density}: 
i) $T_1$ is a tree with nodes $V_1 =  \lbrace 1, \dots , d \rbrace$ and edges $E_1$ ii) For $m \geq 2$, $T_m$ is a tree with nodes $V_m = E_{m - 1}$ and edges  $E_m$ iii) Whenever two nodes in $T_m+1$ are joined by an edge, the corresponding edges in $T_m$ must share a common node.
The corresponding tree sequence is the \emph{structure} of the vine.
Each edge $e$ is associated to a bivariate copula $c_{j_e, k_e | D_e}$, with the set $D_e \in \left\{1, \cdots, d \right\}$ and the indices $j_e, k_e  \in \left\{1, \cdots, d \right\}$ forming respectively its \emph{conditioning set} and the \emph{conditioned set}. 
Finally, the joint copula density can be written as the product of all pair-copula densities $	c(u_1, \cdots, u_d) = \prod_{m=1}^{d-1} \prod_{e \in E_m} c_{j_e, k_e | D_e}(u_{j_e| D_e}, u_{k_e| D_e})$
where $u_{j_e | D_e} = \mathbb{P}\left[ U_{j_e} \leq u_{j_e} \mid  \bm{U}_{D_e} = \bm{u}_{D_e} \right] $
and similarly for $u_{j_e | D_e}$, with $\bm{U}_{D_e} = \bm{u}_{D_e} $ understood as component-wise equality for all components of $(U_1, \dots, U_d)$ and $(u_1, \dots, u_d)$ included in the conditioning set $D_e$.

For self-contained manuscript, we borrow from \cite{tagasovska2019copulas}, a full example of an R-vine for a 5-dimensional density.

\begin{figure}[h]
\centering
\includegraphics[width=0.7\textwidth]{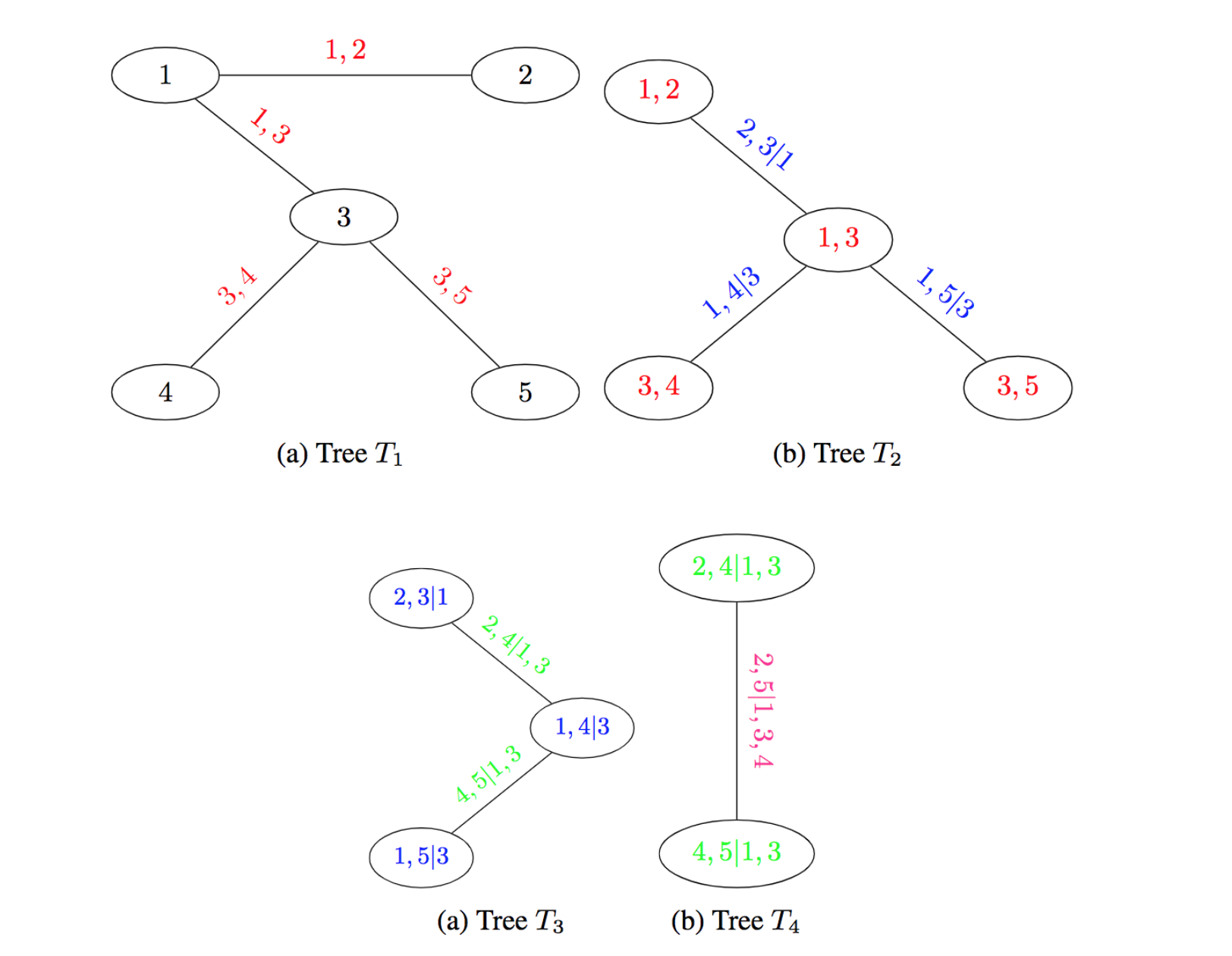}
\label{fig:rvineex4}
 
\caption{A vine tree sequence: the numbers represent the variables, $x,y$ the bivariate distribution of $x$ and $y$, and $x,y|z$ the bivariate distribution of $x$ and $y$ conditional on $z$. Each edge corresponds to a bivariate pair-copula in the PCC.}
	\label{sec:gampcc:RVine_fig}
\end{figure}

\begin{Example}\label{sec:gampcc:vine_ex}
	The density of a PCC corresponding to the tree sequence in \autoref{sec:gampcc:RVine_fig} is
  \begin{align}
	c &= {\color{red}c_{1,2}\,  c_{1,3}\, c_{3,4}\, c_{3,5}}\, {\color{blue} c_{2,3|1}\, c_{1,4|3}\, c_{1,5|3}} {\color{green} c_{2,4|1,3}\, c_{4,5|1,3}}\, {\color{magenta} c_{2,5|1,3,4}},
	\end{align}
	where the colors correspond to the edges {\color{red} $E_1$}, {\color{blue} $E_2$}, {\color{green} $E_3$}, {\color{magenta} $E_4$}.
\end{Example}

\subsection{Details of the toy experiment in \autoref{fig:motivating_example}}
The toy example in \autoref{fig:motivating_example} is generated as: $Y_{train} = X_1 + X_2 + X_3 + \epsilon$, with $X1 = \mathcal{U}[-2\pi, 2\pi]$, $X_2 = sin(2X_1)$, $X_3 \sim sin(X_1^2)$, with $\epsilon \sim \mathcal{N}(0, 0.2)$. For the test data $Y_{train} = X_1 + X_2 + X_3 + x_1\epsilon$, with $X1 \sim \mathcal{U}[-4\pi, 5\pi]$, $X_2 = sin(X_1)$, $X_3 \sim sin(X_1^2)$, with $\epsilon \sim \mathcal{N}(0, 0.5)$. 
Furthermore, the train data consists of 280 samples, and the test data of 100. The network was trained for 100 epochs for the three presented cases, using Adam optimizer with default PyTorch parameters. We use $S$$=$$30$ to get the confidence intervals and $\tau_{low} = 0.025$ and $\tau_{high} = 0.975$ for the prediciton intervals.

\subsection{Details BiLSTM implementation - Datalakes}

Pre-RNN fully connected layers: 1 layer $\times$ 32 units, LeakyReLU ($\alpha=0.2$)
\\RNN layers: 3 LSTM layers $\times$ 32 units each and bidirectional ($\times$ 2 )
\\Dimensionality of last hidden layer: 64
\\Train and test data dimensions: 214,689 and 186,129 samples of 18 dimensions
\\Dropout level: 0.3 
\\Rough estimate of train time: 6 days
\\VC fit time: 2634.29 sec; 
\\VC inference time per data point: 2.1 sec
\\number of VC bootstraps: 15 
\\Framework: Tensorflow 2.4.1

Due to the longer training times of the BiLSTM model and limited time and resources, we were not able to obtain standard variations of the results.
Same reasoning goes for not recommending and including ensembles.
\subsection{Details DenseNet implementation - AirBnB}

The architecture used for the main DenseNet is the same that the proposed in \cite{brando2019umal}. Additionally, the different models proposed in this article satisfies the following parameters and results:

Number of layers: $6$ dense layers
\\
Number of neurons per layer: $120, 120, 60, 60, 10$ and $1$
\\
Training time: $30.4$ secs
\\
Activation types: ReLU activation for hidden layers
\\
Dimensionality of last hidden layer used by the VC: 10
\\
Training , validation and test data dimensions: $29078$, $3634$ and $3633$ respectively.
\\
\\
VC estimates train time: $1861.09 \pm 115.91$ secs.
\\
Time of the VC predicting each point: $1.53 \pm 3 \cdot 10^{-3}$ secs
\\
Number of VC bootstraps: $10$
\\
Framework: Tensorflow 2.3


\end{document}